\documentclass{article}

\usepackage[final]{neurips_2023}
\usepackage[dvipsnames]{xcolor}

\usepackage[utf8]{inputenc} 
\usepackage[T1]{fontenc}    
\usepackage{hyperref}       
\usepackage{url}            
\usepackage{booktabs}       
\usepackage{amsfonts}       
\usepackage{nicefrac}       
\usepackage{microtype}      
\usepackage{xcolor}         

\usepackage{times}
\usepackage{epsfig}
\usepackage{graphicx}
\usepackage{amsmath}
\usepackage{amssymb}
\usepackage{multirow}
\usepackage{tikz}

\usepackage{wrapfig}

\def\R{\mathbb{R}}

\def\x{ {\bf x} }

\def\F{\mathcal{F}}

\def\our{MultiPlaneNeRF}
\def\ourGAN{MultiPlaneGAN}

\title{\our: Neural Radiance Field with Non-Trainable Representation}

\author{
  Dominik Zimny\thanks{Equal contribution}\\
Jagiellonian University
  \And
  Artur Kasymov$^*$\\
Jagiellonian University
  \And
  Adam Kania\\
Jagiellonian University
  \And
  Jacek Tabor\\
Jagiellonian University
  \And
  Maciej Zieba\\
University of Science \\
and Technology Wrocła
  \And
  Marcin Mazur\\
Jagiellonian University
  \And
  Przemyslaw Spurek\\
Jagiellonian University
\thanks{\texttt{przemyslaw.spurek@uj.edu.pl}}
}

\begin{document}

\maketitle

\begin{abstract}

NeRF is a popular model that efficiently represents 3D objects from 2D images. However, vanilla NeRF has some important limitations. NeRF must be trained on each object separately. The training time is long since we encode the object's shape and color in neural network weights. Moreover, NeRF does not generalize well to unseen data. In this paper, we present MultiPlaneNeRF -- a model that simultaneously solves the above problems. Our model works directly on 2D images. We project 3D points on 2D images to produce non-trainable representations. The projection step is not parametrized and a very shallow decoder can efficiently process the representation. Furthermore, we can train MultiPlaneNeRF on a large data set and force our implicit decoder to generalize across many objects. Consequently, we can only replace the 2D images (without additional training) to produce a NeRF representation of the new object. In the experimental section, we demonstrate that MultiPlaneNeRF achieves results comparable to state-of-the-art models for synthesizing new views and has generalization properties. Additionally, MultiPlane decoder can be used as a component in large generative models like GANs.           
\end{abstract}

\section{Introduction}

Neural Radiance Fields (NeRFs)~\cite{mildenhall2020nerf} enable synthesizing novel views of complex scenes from a few 2D images with known camera positions. This neural network model can reproduce high-quality scenes from previously invisible viewpoints based on the relationship between these basic images and computer graphics principles such as radiation tracking.
NeRFs~\cite{mildenhall2020nerf} represent a scene using a fully connected architecture. NeRF takes a 5D coordinate as the input: spatial location and camera positions. On outputs, we obtain color and volume density.
The loss of NeRF is inspired by classical volume rendering~\cite{kajiya1984ray}.  We render the color of all rays passing through the scene. 
In practice, the shape and colors of the 3D object are encoded in neural network weights. 

\begin{figure}[!h]
    \begin{center}
    \begin{tikzpicture}[scale=0.5]
    \node[inner sep=0pt] (russell) at (0,0)
    {\includegraphics[width=0.5\textwidth]{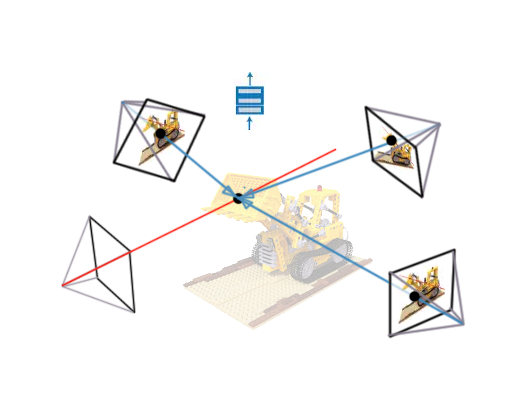} };
    \node[text width=3.5cm] at (2,1.8) {$ Z_{(x,y,z)}$};
    \node[text width=3.5cm] at (2.1,4.7) { RGB $\sigma$};
    \end{tikzpicture}
    \caption{In \our{} approach, we divided 2D training images into two parts. The first one builds a 2D representation and is used as input to a small implicit decoder. The second part is used as a vanilla NeRF training data set. The representation of a 3D object containing $n$ 2D images is part of the architecture. The implicit decoder takes the coordinates of the 3D point $(x,y,z)$ and applies projection on the given 2D images. Then the aggregate information of the projected pixel $Z_{(x,y,z)} \in \R^{5k}$ is used to predict the color RGB and the volume density $\sigma$.   } 
    \label{fig:architecture}
    \end{center}
\end{figure}

NeRF architecture produces extremely sharp renders of new views of a static scene. Unfortunately, such a model has a few important limitations. NeRF must be trained on each object separately, and it does not generalize to unseen data.
The training time is long since we encode the object's shape in neural network weights.

\begin{figure}[!h]
    \begin{center}
    \begin{tikzpicture}[scale=0.5]
    \node[inner sep=0pt] (russell) at (0,0)
    {\includegraphics[width=0.49\textwidth]{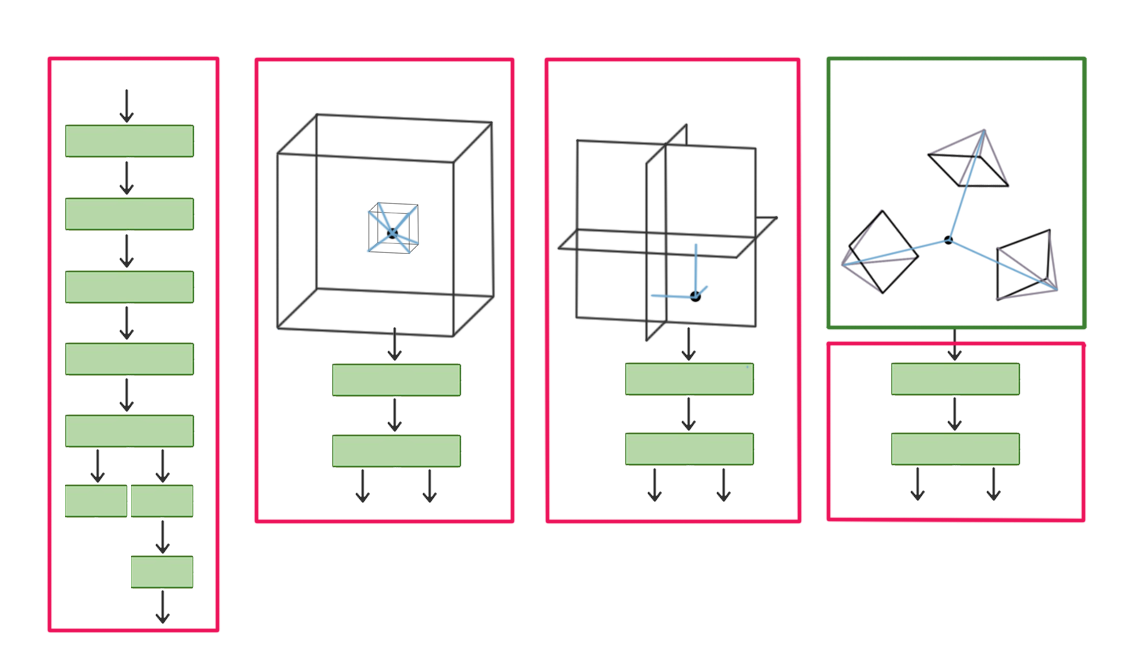} };
    \node[text width=3.5cm] at (-4.5,5.2) {(a) NeRF };
    \node[text width=3.5cm] at (-1.1,5.2) {(b) Voxels };
    \node[text width=3.5cm] at (3.5,5.2) { (c) TriPlane };
    \node[text width=3.5cm] at (7.8,5.2) { (d) MultiPlane };
    \node[text width=7cm] at (3.,-6.5) { \color{purple} Trainable parameters };
    \draw [very thick, purple](-6.0,-6.5) -- (-5.0,-6.5);
    
    \draw [very thick, teal](-6.0,-5.5) -- (-5.0,-5.5);
    \node[text width=7cm] at (3.,-5.5) { \color{teal} Non-Trainable parameters (input images) };
    \end{tikzpicture}
    \caption{Neural implicit representations use fully connected layers with position encoding to represent a scene (a). Explicit voxel grids or hybrid variants using small implicit decoders are fast but scale poorly with resolution (b). Hybrid explicit-implicit TriPlane representation is fast and well scale, but we must train its parameters (c). In Hybrid explicit-implicit MultiPlane representation, we use existing images as a representation and use a small implicit decoder to aggregate information. By ref color, we marked trainable parameters of respected models.  
    } 
    \label{fig:merf_appro}
    \end{center}
\end{figure}
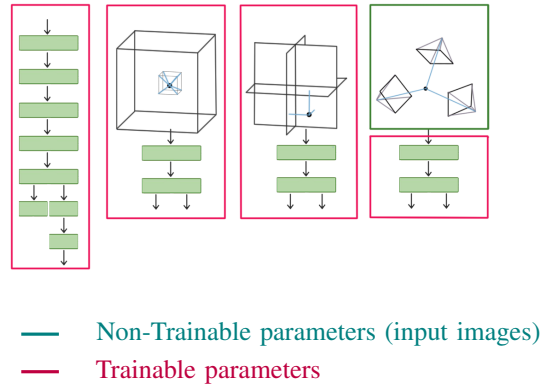

Therefore, several modifications of NeRF have appeared to solve the above problems. 
In practice, trainable voxel-bounded implicit fields~\cite {liu2020neural,fridovich2022plenoxels,sun2022direct,muller2022instant} can be used as a representation of 3D objects. Instead of encoding the 3D structure in the weights of the deep model, we train voxels and a small implicit decoder (MLP) to predict RGB colors and volume density.
Such a solution reduces the training and inference times, see Fig.~\ref{fig:merf_appro}.
Alternatively, we can use a TriPlane concept described in \cite{chan2022efficient,chen2022tensorf}, based on training orthogonal planes to represent 3D objects. Similar to voxel-type NeRFs, TriPlane-based models use a small MLP (implicit decoder) to aggregate information and predict RGB colors and volume density. The planes are trained together with the implicit decoder.

Voxel and plane-based representations reduce the computational time and increase the model's accuracy. 
But the above models do not generalize well to unseen data.
To solve such a problem, we can use existing 2D images and an extensive network to extract information~\cite{chen2021mvsnerf,wang2021ibrnet}. Thanks to the trainable feature extractor, we can train models on a large number of various objects.    
But architecture becomes vast and training time increases drastically.    

In this paper, we present \our{}\footnote{The source code is available at: \url{https://github.com/gmum/MultiPlaneNeRF.}} -- new NeRF model with easy-to-train small architecture, which has generalization properties. Our model works directly on 2D images. We project 3D points on 2D images to produce non-trainable representations. The projection step is not parametrized, and a very shallow decoder can efficiently process the representation.
In \our{}, we split the initial set of 2D training images into two subsets. The first one is used to build a 2D representation and further used as input to a small implicit decoder, see Fig.~\ref{fig:merf_appro}. The second one is utilized as a training set to calculate the weights of the decoder.  
Furthermore, we can train \our{} on a large data set and force our implicit decoder to generalize across many objects. Consequently, we can only change the 2D image to produce a NeRF representation of the new object.

\our{} decoder can be used not only as a NeRF representation of a 3D object but also similarly to the TriPlane decoder as a component in a large generative model such as GAN~\cite{chan2022efficient}.

To summarize, the contributions of our work are the following:
\begin{itemize}
    \item We propose a new method dubbed~\our{} which uses non-trainable representations of 3D objects.
    \item \our{}  achieves comparable results to state-of-the-art models for synthesizing new views and can generalize to unseen objects by changing image-based representation without additional training.
    \item We propose \ourGAN{} -- a GAN-based generative model that uses a MultiPlane decoder as an interpretable representation of the 3D objects.
\end{itemize}

\section{Related Works}

3D objects can be represented by using many different approaches, including voxel grids~\cite{choy20163d}, octrees \cite{hane2017hierarchical}, multi-view images \cite{arsalan2017synthesizing,LIU2022108774}, point clouds \cite{achlioptas2018learning,shu2022wasserstein,yang2022continuous}, geometry
images \cite{sinha2016deep}, deformable meshes \cite{girdhar2016learning,li2017learning},
and part-based structural graphs \cite{li2017grass}.

The above representations are discreet, which causes some problems in real-life applications. In contrast to such apprehension, NeRF~\cite{mildenhall2020nerf} represents a scene using a fully-connected architecture. NeRF and many generalizations \cite{barron2021mip,barron2022mip,liu2020neural,niemeyer2022regnerf,roessle2022dense,tancik2022block,verbin2022ref}  synthesize novel views of a static scene using differentiable volumetric rendering.

One of the largest limitations is training time. To solve such problems in \cite{fridovich2022plenoxels}, authors propose Plenoxels, a method that uses
a sparse voxel grid storing density and spherical harmonics coefficients at each node. The final color is the composition of tri-linearly interpolated values of each voxel. In DVGO~\cite{sun2022direct} also optimize voxel grids of features for fast radiance field reconstruction.
In \cite{muller2022instant}, authors use a similar approach, but space is divided into an independent multilevel grid.
In \cite{chen2022tensorf}, authors represent a 3D object as an orthogonal tensor component. A small MLP network, which uses orthogonal projection on tensors, obtains the final color and density. 
There exist some methods which use additional information to Nerf, like depth maps or point clouds
\cite{azinovic2022neural,deng2022depth,roessle2022dense,wei2021nerfingmvs}.

Many approaches are dedicated to training models on a few existing views. Most of the method uses a large feature extractor trained on many different objects to allow generalization properties. 
In \cite{chen2021mvsnerf}, authors build a cost volume at the reference view by
warping 2D neural features onto multiple sweeping planes. Then authors use a 3D CNN to aggregate information.
In \cite{wang2021ibrnet}, authors use a large feature extractor and a new projection strategy to build a 3D object representation that can generalize across different objects. In the end, a small network called a ray transformer aggregates information. In pixelNeRF \cite{yu2021pixelnerf}, the convolutions layer transfers input images to produce a representation for NeRF based model.

EG3D~\cite{chan2022efficient} uses a tri-plane representation for 3D GANs; their representation is similar to our TensoRF~\cite{chen2022tensorf} and exists only as a part of large generative models.

The above models solve some of the most critical NeRF limitations. But no model solves all of them simultaneously.
In this paper, we present \our{} -- new NeRF model with easy-to-train small architecture, which has generalization properties.

\section{\our{}:  NeRF with generalization properties}
\label{sec:method}


This section briefly describes the three most popular NeRF representations: vanilla NeRF, voxel NeRF, and TriPlane NeRF. Next, we provide the details about \our{} the novel alternative rendering approach with non-parametric representations (see Fig.~\ref{fig:merf_appro}). 

\paragraph{NeRF representation of 3D objects}

Vanilla NeRF~\cite{mildenhall2020nerf}  is the model for representing complex 3D scenes using neural architectures. NeRF takes a 5D coordinate as input, which includes the spatial location $ \x = (x, y, z)$ and viewing direction ${\bf d} = (\theta, \psi)$ and returns emitted color ${\bf c} = (r, g, b)$ and volume density $\sigma$. 

A vanilla NeRF uses a set of images for training. In such a scenario, we produce many rays traversing through the image and a 3D object represented by a neural network. 
NeRF approximates this 3D object with an MLP network:
$$
\F_{NeRF} (\x , {\bf d}; \Theta ) = ( {\bf c} , \sigma).
$$
The model is parameterized by $\bf \Theta$ and trained to map each input 3D coordinate to its corresponding volume density and directional emitted color.  

The loss of NeRF is inspired by classical volume rendering \cite{kajiya1984ray}.  We render the color of all rays passing through the scene. The volume density $\sigma( \x )$ can be interpreted as the differential probability of a ray. The expected color
$C({\bf r})$ of camera ray ${\bf r}(t) = {\bf o} + t {\bf d} $ (where  ${\bf o}$ is ray origin and ${\bf d}$ is direction) can be computed with an integral, but in practice, it is estimated numerically using stratified sampling. The loss is simply the total squared error between the rendered and true pixel colors:

\begin{equation}
    \mathcal{L} = \sum _{{\bf r} \in R} \| \hat C({\bf r}) - C({\bf r}) \|_2^2,
    \label{eq:cost_general}
\end{equation}

where $R$ is the set of rays in each batch, and $C({\bf r})$, $\hat C ({\bf r})$ are the ground truth and predicted RGB colors for ray {\bf r } respectively.   

The predicted RGB colors $\hat C ({\bf r})$  can be calculated with formula:

\begin{equation}
\hat C({\bf r}) = \sum_{i=1}^{N} T_i (1-\exp(-\sigma_i \delta_i)) {\bf c}_i, \mbox{ where } T_i=\exp \left(-\sum_{j=1}^{i-1} \sigma_i \delta_i \right)
\end{equation}

where $N$ is the number of samples, $\delta_i$ is the distance between adjacent samples, and $\sigma_i$ denotes the opacity of sample $i$. This function for calculating $\hat C({\bf r})$ from the set of $(c_i, \sigma_i)$ values is trivially differentiable.

In practice, we encode the structure of 3D objects into neural network weights. Such architecture is limited by network capacity or difficulty finding accurate intersections of camera rays with the scene geometry. Synthesizing high-resolution imagery from
these representations often require time-consuming optical ray marching.




\begin{figure}
{\small Original render \quad \our{} \quad Original render \quad \our{}  \ Original render \quad \our{} }\\
     {\includegraphics[width=0.15\textwidth]{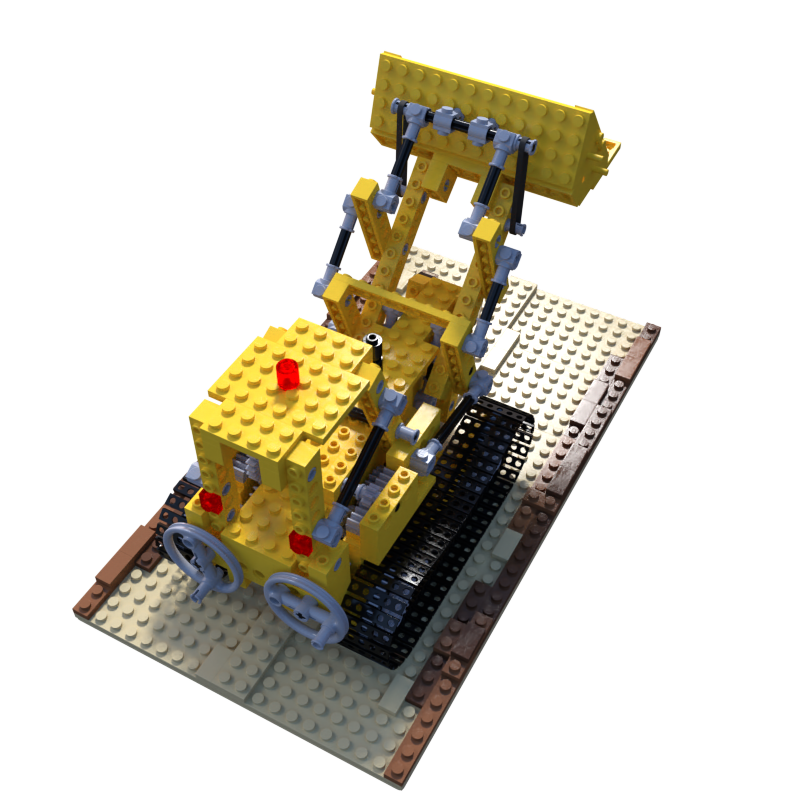} }
    {\includegraphics[width=0.15\textwidth]{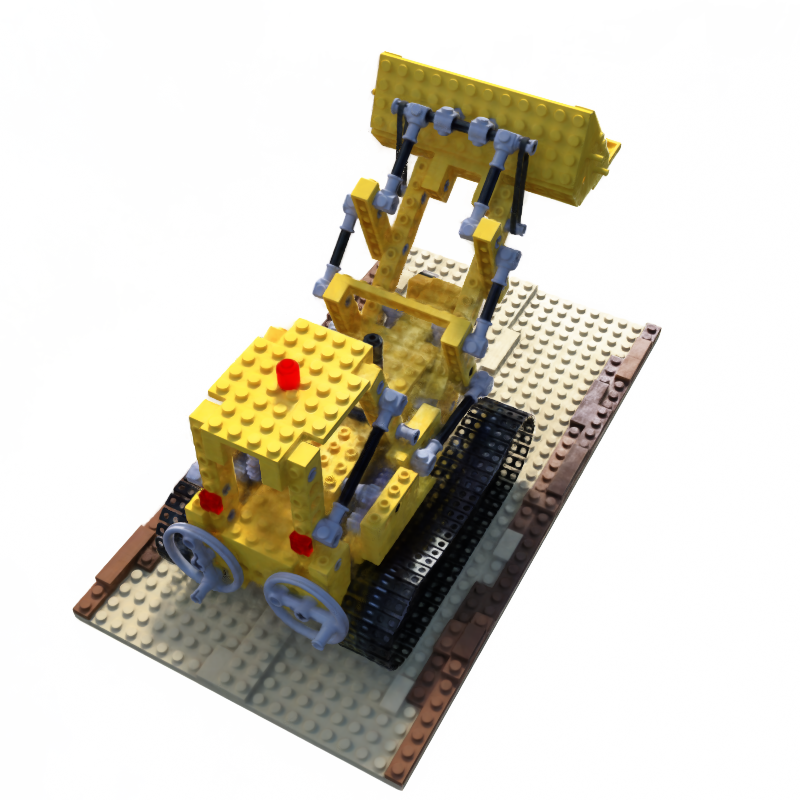} }
    {\includegraphics[width=0.15\textwidth]{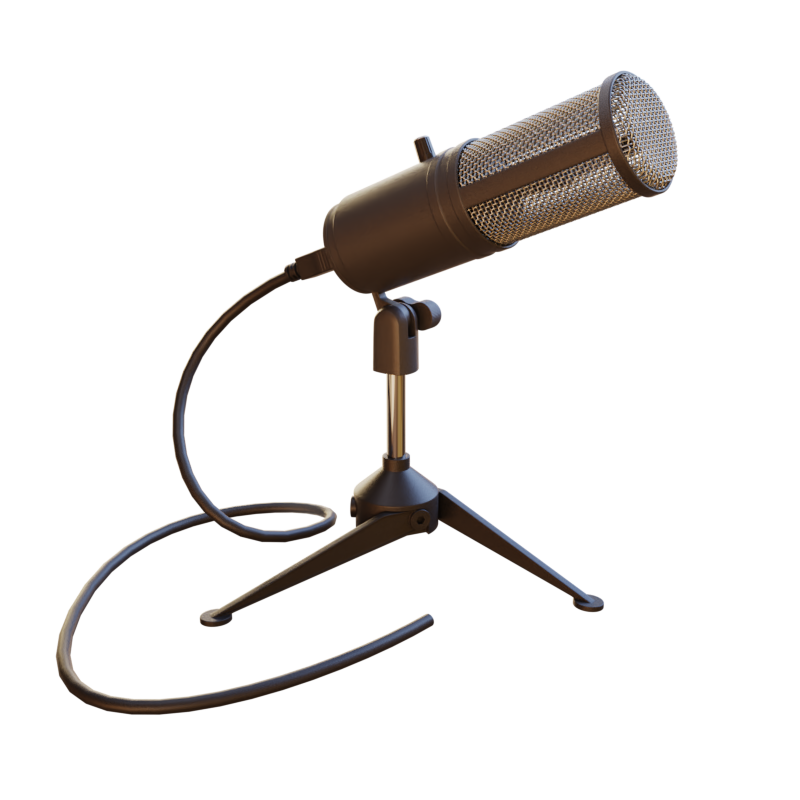} }
    {\includegraphics[width=0.15\textwidth]{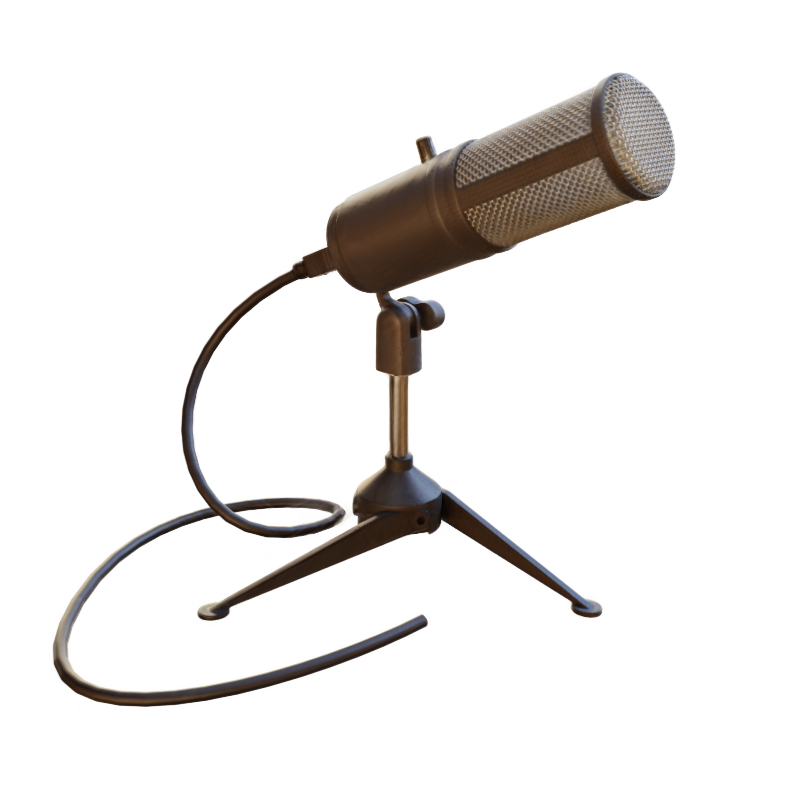} }
    {\includegraphics[width=0.15\textwidth]{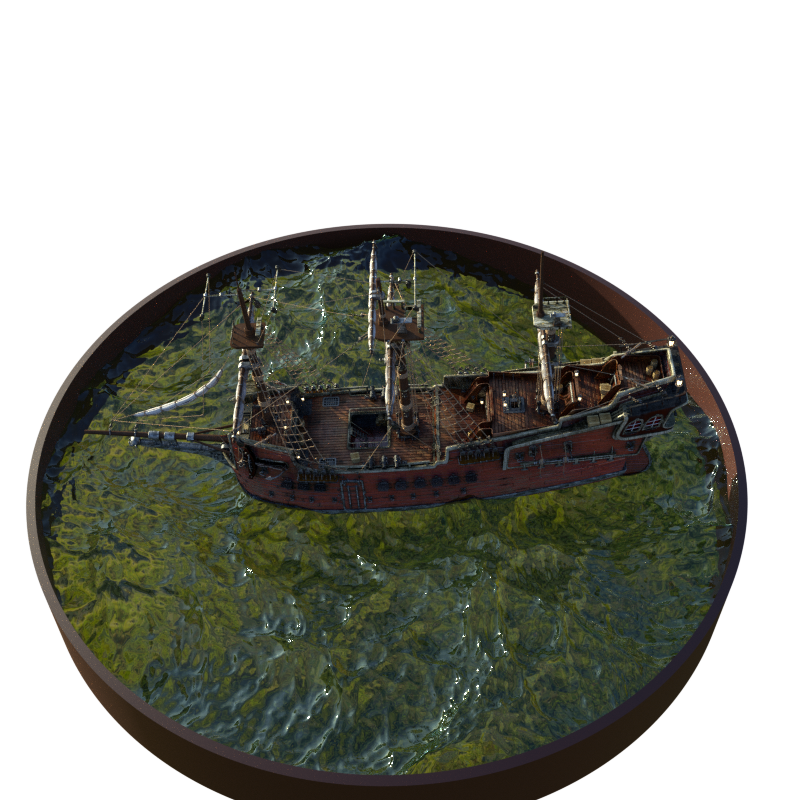} }
    {\includegraphics[width=0.15\textwidth]{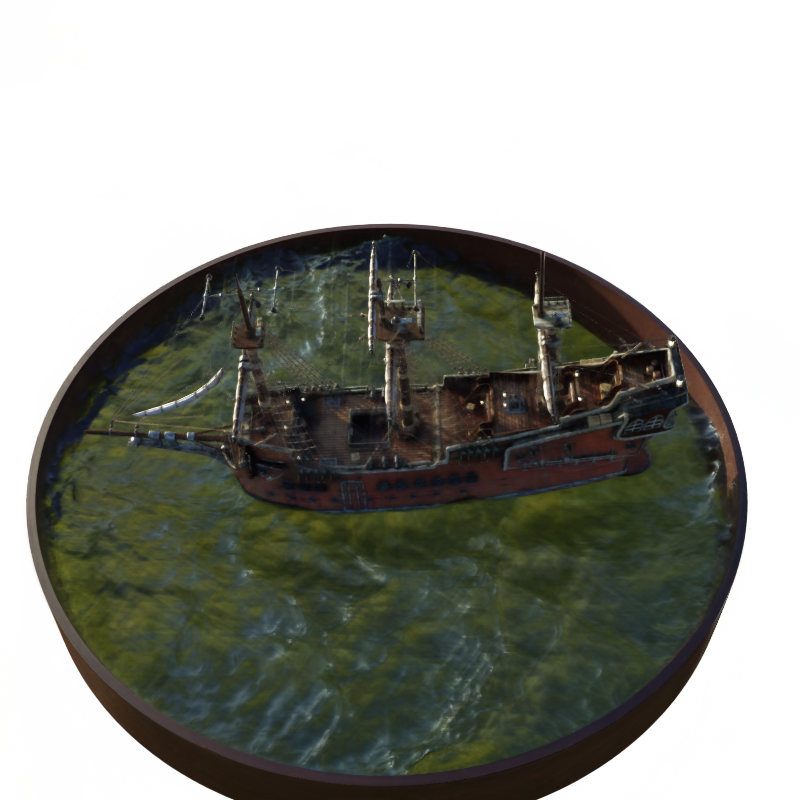} }
    {\includegraphics[width=0.15\textwidth]{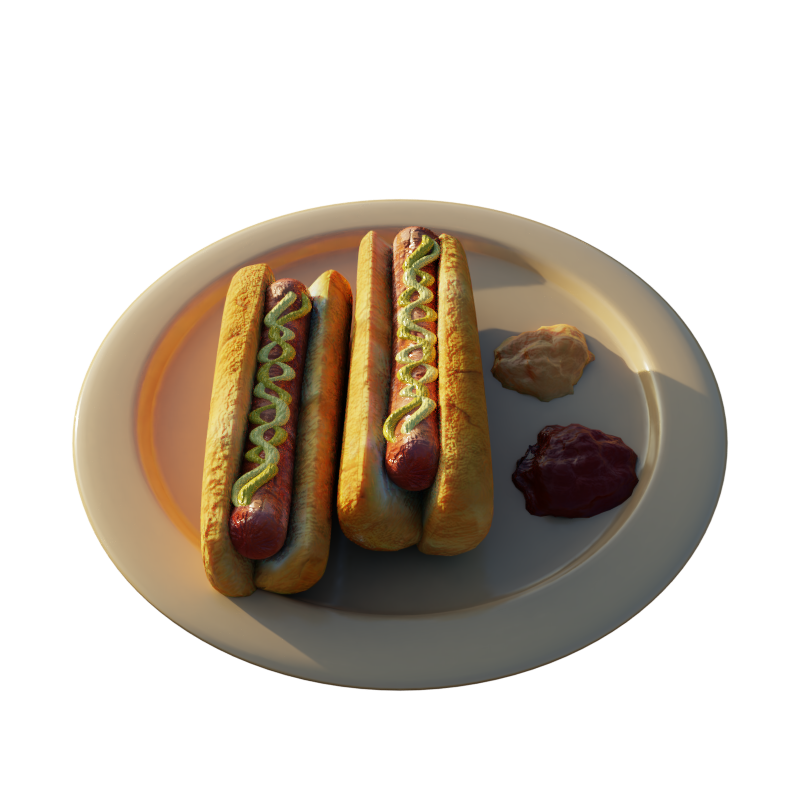} }
    {\includegraphics[width=0.15\textwidth]{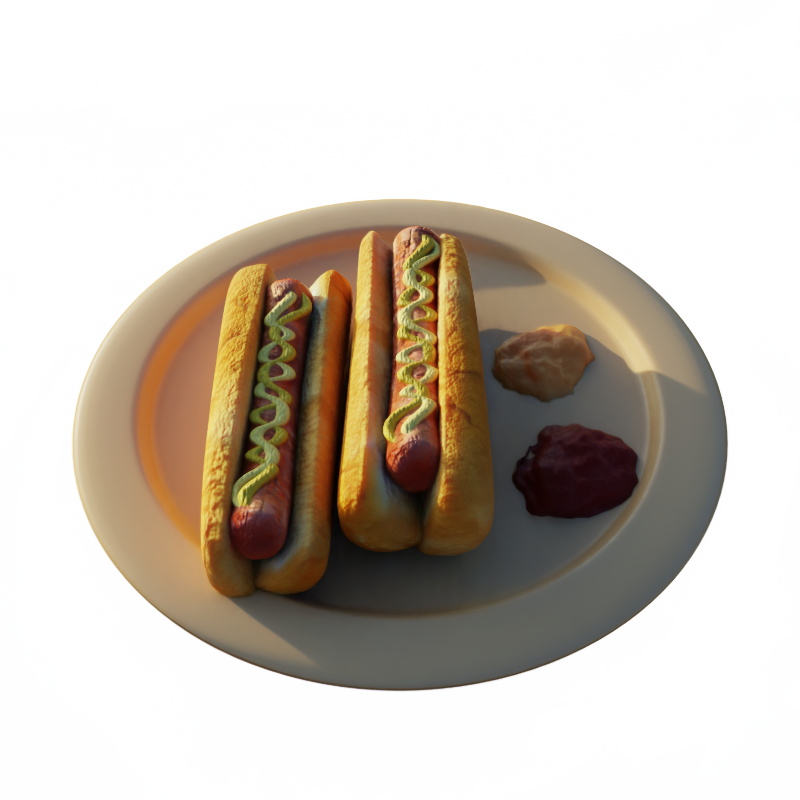}
    {\includegraphics[width=0.15\textwidth]{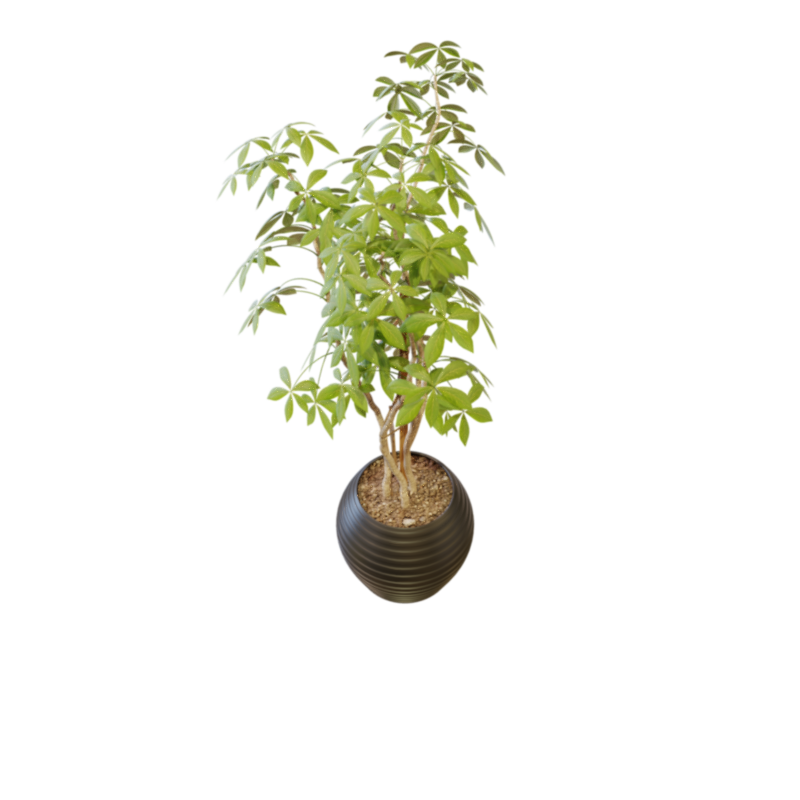} }
    {\includegraphics[width=0.15\textwidth]{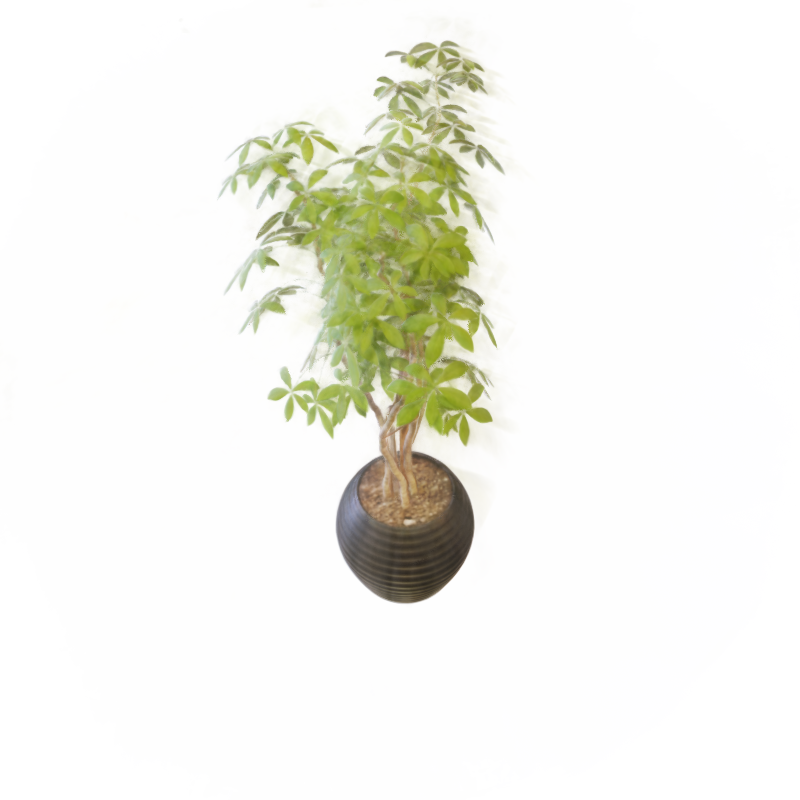} }}
    {\includegraphics[width=0.15\textwidth]{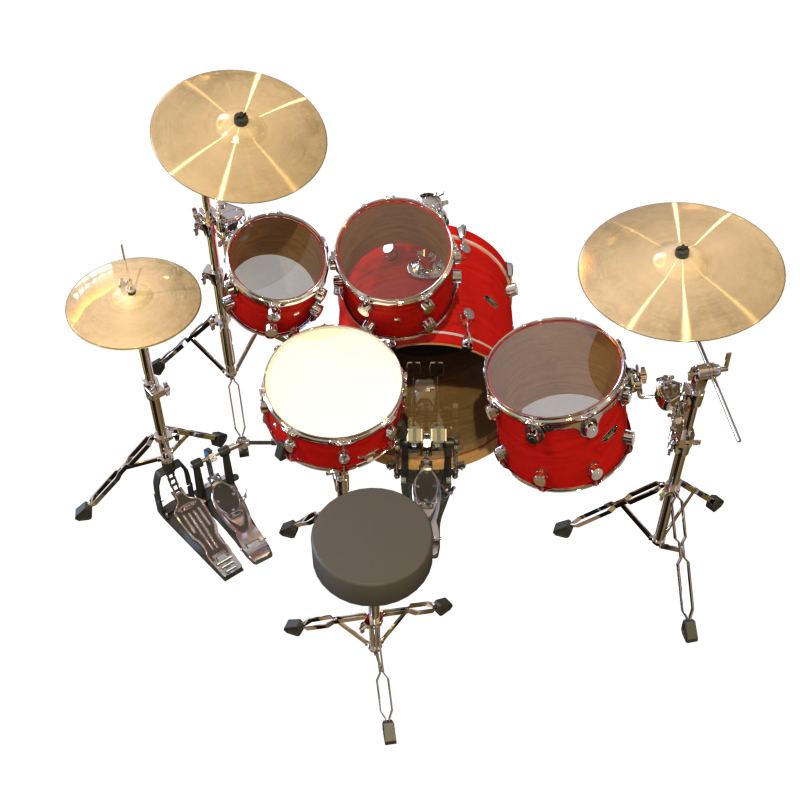} }
    {\includegraphics[width=0.15\textwidth]{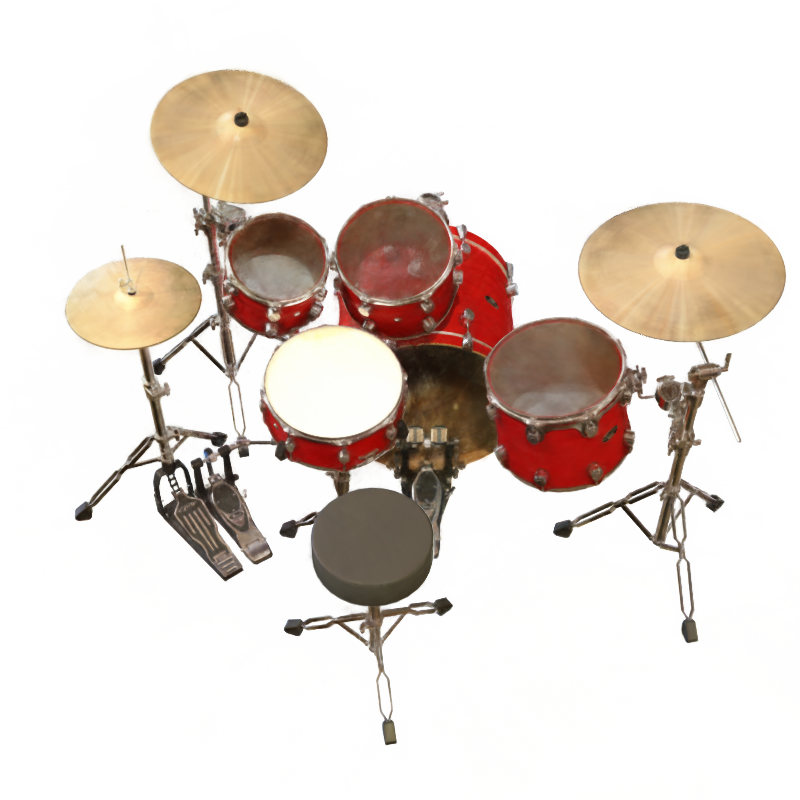} }

\caption{Visualization of renders produce by \our{} on NeRF Synthetic dataset scenes: \textit{Lego}, \textit{Mic}, \textit{Ship}, \textit{Hotdog}, \textit{Drums}, \textit{Ficus}.}
\label{fig:rec_ner}
\end{figure}

\paragraph{Neural Voxel Fields }

Instead of modeling the entire space with a single
implicit function, we can use voxel-bounded implicit fields~\cite{fridovich2022plenoxels,liu2020neural,muller2022instant,sun2022direct}. Specifically, we assign a voxel embedding at each vertex and obtain a new representation. A small implicit decoder can aggregate information from voxel representation  to model RGB colors and volume density, see sub-figure (b) in Fig.~\ref{fig:merf_appro}.

We assume that we have a grid $n \times m$ of voxels, and each of them is represented by trainable embedding $\mathbf{v}_{i,j}$:
$$
V = \{\mathbf{v}_{i,j} \in \R^k \}, \mbox{ where }
 i=1,\ldots,n, \mbox{ and } j = 1. \ldots, m,
 $$
 is a representation of 3D object. In practice, we use a sparse voxel structure, where nodes exist only in non-empty areas of 3D space.

A small implicit decoder aggregates information from such voxel representation of 3D objects to model RGB color and volume density:
$$
\F_{Voxel NeRF}(\x , {\bf d}; \Theta, V ) = ( {\bf c} , \sigma). 
$$
The model is parameterized by $\bf \Theta$ and voxel representation $V$. We train the model to map each input 3D coordinate to its corresponding volume density and directional emitted color.
In practice, for position $\x \in \R^3$, we aggregate information from the six nearest voxels from the grid, see sub-figure (b) in Fig.~\ref{fig:merf_appro}. 
The loss function and rendering procedure are directly taken from vanilla NeRF.

Voxel representation allows faster training, but there is a problem with scaling the solution to a higher resolution.

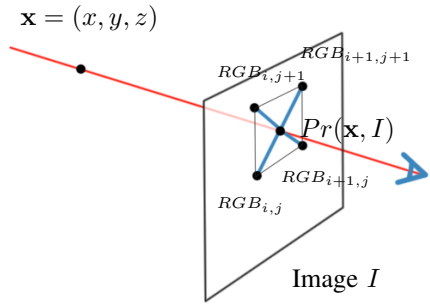
\begin{wrapfigure}{r}{0.5\textwidth}
\vspace{-1cm}
    \begin{center}
    \begin{tikzpicture}[scale=0.5]
    \node[inner sep=0pt] (russell) at (0,0)
    {\includegraphics[width=0.5\textwidth]{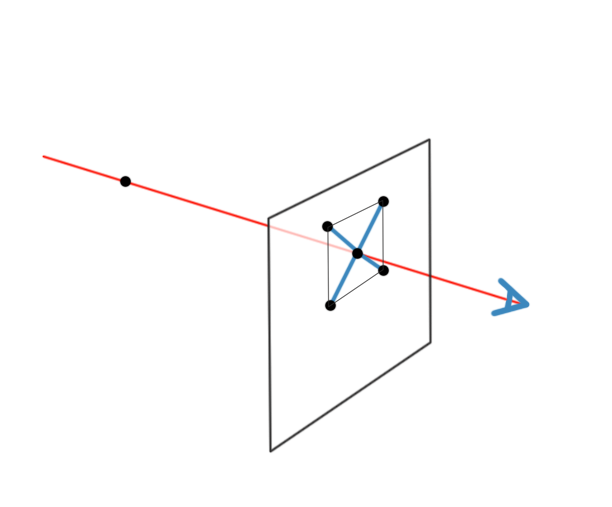} };
    \node[text width=3.5cm] at (5.2,2) {\tiny $RGB_{i+1,j+1}$};
    \node[text width=3.5cm] at (4.7,-1.3) {\tiny $RGB_{i+1,j}$};
    \node[text width=3.5cm] at (3.0,-2.0) {\tiny $RGB_{i,j}$};
    \node[text width=3.5cm] at (3.0, 1.5) {\tiny $RGB_{i,j+1}$};
    \node[text width=3.5cm] at (5,-4) { Image $I$};
    \node[text width=3.5cm] at (-2.2,3) { $\x = (x,y,z)$};
    \node[text width=3.5cm] at (5.2,0) { $ Pr(\x,I)$};    
    \end{tikzpicture}
    \caption{ For input 3D point $\x=(x,y,z)$ we apply its projection on image $I$ and obtain 2D coordinate $Pr(\x,I)$. Then we use linear interpolation of colors from four closes pixel colors $RGB_{i,j}$, $RGB_{i+1,j}$, $RGB_{i,j+1}$, $RGB_{i+1,j+1}$ to the estimated color $RGB_{Pr(\x,I)}$ in position $Pr(\x,I)$. Position and colors $[ RGB_{Pr(\x,I)}, Pr(\x,I) ]$ are input to implicit decoder.   } 
    \label{fig:projection}
    \end{center}
\end{wrapfigure}

\paragraph{ TriPlane NeRF }

To solve the above problems, we can use TriPlane~\cite{chan2022efficient} 3D representation that is both efficient and expressive, see sub-figure (c) in Fig.~\ref{fig:merf_appro}.

In the TriPlane formulation, we align our explicit features along three axis-aligned orthogonal feature planes $P = (P_{xy}, P_{xz}, P_{yz})$, each with a resolution of $N \times N \times C$ (Fig..~\ref{fig:merf_appro} (c)) with $N$ is spatial resolution and $C$ the number of channels. We query any 3D position $x \in \R^3$ by projecting it onto each of the three feature planes, retrieving the corresponding feature vector $(F_{xy}, F_{xz}, F_{yz})$ via bilinear interpolation, and aggregating the three feature vectors via summation.
An additional lightweight decoder network implemented as a
small MLP interprets the aggregated 3D features 
$$
\F_{TriPlane NeRF}(\x , {\bf d}; \Theta, P ) = ( {\bf c} , \sigma). 
$$
Similarly to Voxel NeRF, we use a vanilla NeRF cost function and rendering procedure. 

The primary advantage of this hybrid representation is keeping the decoder small and using explicit feature representation. Plane representation can be calculated with gradient-based optimization or produced by convolutions neural networks like StyleGAN or diffusion model.

\paragraph{ MultiPlane NeRF }

MultiPlane NeRF is the next step toward efficient and expressive representation of 3D objects. Unlike voxel and TriPlane models, this approach does not rely on a trainable representation scheme. Instead, it employs pre-existing 2D images as a planar representation, similar to the TriPlane model. By using a fixed set of $n$ 2D images, MultiPlane NeRF enables the training of a small implicit decoder. This approach presents a practical solution for achieving efficient and expressive 3D object representation. 

Our framework uses the projection of point $\x \in \R^3$ on $n$ fixed 2D images. We stay in NeRF framework, which reconstructs real scenes with the camera-to-world scenario. 
In NeRF rendering procedure, we have ray and 2D images, and our goal is to model 3D coordinates. 
Using the same transformation, we can reverse the process to obtain  world-to-camera. From 3D coordinates $\x = (x,y,z)$, we obtain 2D coordinates on image $I$.
In practice we apply projection of $\x = (x,y,z)$ on image $I$:
$$
Pr(\x, I) = \mathbf{z}_{I} \in \R^2,
$$
where $\mathbf{z}_{I}$ represents the point in $\R^2$ created by projecting $\x$ on image $I$.

\begin{table*}[]
\begin{center}
    \begin{tabular}{l|llllllll}
    \multicolumn{9}{c}{PSNR $\uparrow$} \\
         & Chair & Drums & Ficus & Hotdog & Lego & Materials & Mic & Ship \\ \hline
    SRN  & 26.96 & 17.18 & 20.73 & 26.81 & 20.85 & 18.09 & 26.85 & 20.60 \\
    NV   & 28.33 & 22.58 & 24.79 & 30.71 & 26.08 & 24.22 & 27.78 & 23.93 \\
    LLFF & 28.72 & 21.13 & 21.79 & 31.41 & 24.54 & 20.72 & 27.48 & 23.22  \\
    NeRF & 33.00 & 25.01 & 30.13 & \bf 36.18 & 32.54 & 29.62 & 32.91 & 28.65  \\
    NSFV & \bf 33.19 & \bf 25.18 & \bf 32.29 & 34.27 & \bf 32.68 & 27.93 & \bf 37.14 & \bf 31.23 \\
    \our{} & 32.81 & 24.28 & 28.22 & 35.75 & 28.49 & \bf 30.80 & 32.70 & 27.39     \\ \hline
    \multicolumn{9}{c}{SSIM $\uparrow$} \\
         & Chair & Drums & Ficus & Hotdog & Lego & Materials & Mic & Ship \\ \hline
   SRN & 0.910 & 0.766 & 0.809 & 0.947 & 0.808 & 0.757 & 0.923 & 0.849 \\
NV & 0.916 & 0.873 & 0.880 & 0.946 & 0.888 & 0.784 & 0.944 & 0.910 \\
NeRF & 0.967 & 0.925 & \bf 0.961 & 0.980 & 0.949 & 0.856 & 0.974 & 0.964 \\
NSFV & 0.968 & \bf 0.931 & 0.960 & \bf 0.987 & \bf 0.973 & 0.854 & 0.980 & \bf 0.973 \\
    \our{} & \bf 0.972 & 0.921  & 0.950 & 0.974 & 0.953 & \bf 0.940  & \bf 0.983 & 0.865    \\ \hline
    \multicolumn{9}{c}{LPIPS $\downarrow$ } \\
         & Chair & Drums & Ficus & Hotdog & Lego & Materials & Mic & Ship \\ \hline
SRN & 0.106 & 0.267 & 0.200 & 0.063 & 0.174 & 0.299 & 0.100 & 0.149 \\
NV & 0.109 & 0.214 & 0.175 & 0.107 & 0.130 & 0.276 & 0.109 & 0.162 \\
NeRF & 0.046 & 0.091 & 0.050 & 0.028 & 0.063 & 0.206 & 0.121 & 0.044 \\
NSVF & 0.043 & \bf 0.069 & \bf 0.029 & \bf 0.010 & \bf 0.021 & 0.162 & 0.025 & \bf 0.017 \\
    \our{} & \bf 0.026 & 0.071 & 0.047 & 0.033 & 0.047 & \bf 0.055 & \bf 0.014 & 0.153    \\ \hline
    \end{tabular}
    \end{center}
    \caption{Numerical comparison of our model and classical SRN~\cite{sitzmann2019scene}, NV~\cite{lombardi2019neural},
LLFF~\cite{mildenhall2019local}, and original  NeRF~\cite{mildenhall2020nerf}, and Voxel based model NSFV~\cite{liu2020neural} on rendering task. Our model obtains comparable results with the original NeRF and voxel-based method using less number of trainable parameters. }
    \label{tab:nerfob}
\end{table*} 

\begin{figure*}[]
\begin{center}
\includegraphics[width=0.9\textwidth]{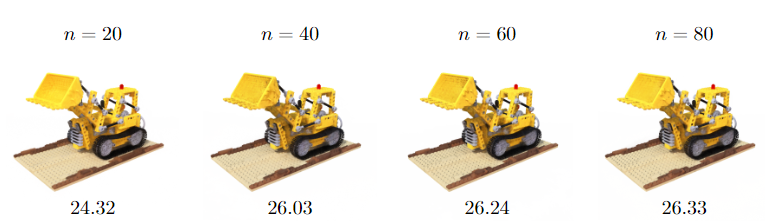}

\end{center}

\caption{Visualization of PSNR metric concerning the number of images used from object representations. We train \our{} for 40k epochs. As we can see, our model obtains better results when we increase the number of images in representations.}
\label{fig:abl1}
\end{figure*}

\begin{figure*}[]
\begin{center}
\includegraphics[width=0.9\textwidth]{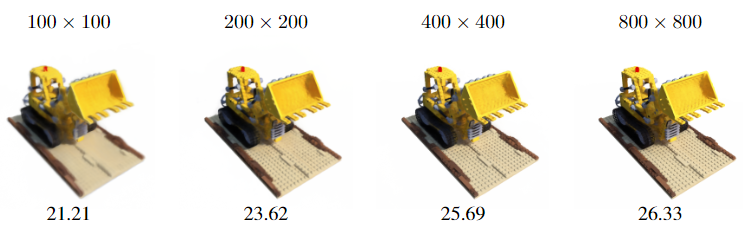}
\end{center}
\caption{Visualization of PSNR metric concerning the resolution of images used from object representations. We train \our{} for 40k epochs on different image resolution and then render the final image with $800\times800$ size. As we can see, our model obtains better results when trained images and expected image resolution match.}
\label{fig:abl2}
\end{figure*}

\begin{figure}
\begin{center}
{\small Original render \quad auto-encoder \quad \our{}  \quad Original render \quad Auto-encoder \quad \our{}  }\\
\includegraphics[width=0.95\textwidth]{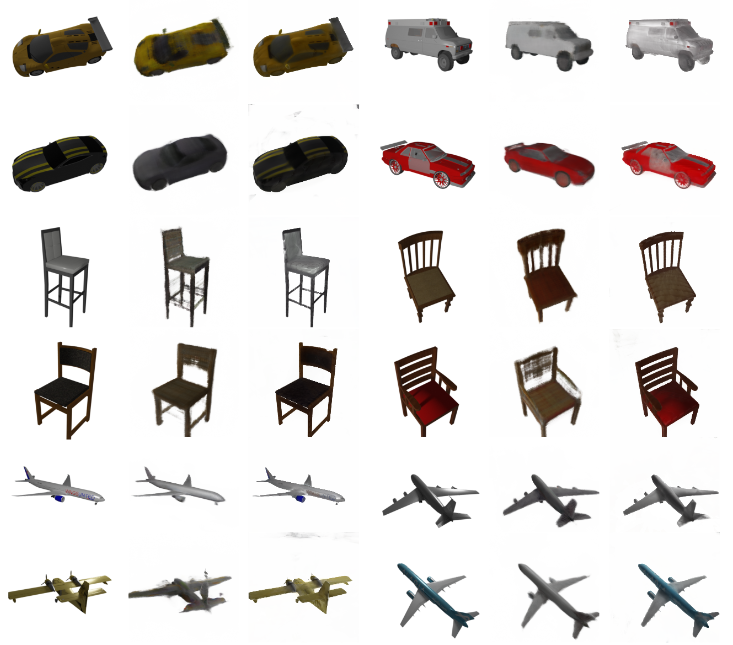}
\end{center}
\caption{Reconstructions of objects from test set produced by auto-encoder Points2NeRF and \our{}. As one can observe, \our{} can produce slightly better renders on the test set than the auto-encoder architecture.}
\label{fig:generalization}
\end{figure}

In \our{} approach, we assume that we have $n$ training images $I_i$ for $i =1, \ldots, n$, which we understand as a non-trainable representation of a 3D object, each with a resolution of $N \times N \times 3$ (Fig..~\ref{fig:merf_appro} (d)).
Such images will be part of the architecture and must be stored with weights of \our{} model.

Our neural network takes 3D point coordinates $\x=(x,y,z)$ and applies projection on each of given 2D images: 
$$
[Pr(\x, I_1), \ldots, Pr(\x, I_n) ] = [\mathbf{z}_{I_1}, \ldots, \mathbf{z}_{I_k}] \in \R^{2n}.
$$
Then we add RGB color from 2D images. Since rays do not cross the coordinates of pixels, we use linear interpolation to obtain color in point $\x_{I_i}$ on image $I_i$, see Fig.~\ref{fig:projection}.
In consequence, we obtain input to NeRF implicit decoder:
$$
Z_{(x,y,z)} =  \left[ I_1[\mathbf{z}_{I_1}] , \mathbf{z}_{I_1} , \ldots, I_1[\mathbf{z}] , \mathbf{z}_{I_n} \right] \in \R^{5 n},
$$
where $I_i[\mathbf{z}_{I_i}] \in \R^3$ is RGB color of position $\mathbf{z}_{I_i}$  on the image $I_i$. 
Our implicit decoder $F_{MultiPlane NeRF}$ aggregates color and positions to produce  RGB colors and the volume density $\sigma$ 
$$
\F_{MultiPlane NeRF}(\x , {\bf d},I_1, \dots, I_n; \Theta) = ( {\bf c} , \sigma). 
$$

The loss function and rendering procedure are directly taken from vanilla NeRF. 


\begin{wraptable}{r}{7.5cm}
\begin{center}
\vspace{-0.6cm}
    \begin{tabular}{ccc}
    & \our{} &  Auto-encoder \\[0.1ex] 
     \hline
     planes (train) & \bf 25.28  &  24.83   \\[0.1ex] 
     planes (test) & \bf 24.26  & 14.18    \\[0.1ex] 
     cars (train) &  26.21 & \bf 28.14  \\[0.1ex]
     cars  (test) & \bf 24.79 & 20.86    \\[0.1ex]
     chairs (train) & \bf 25.28 &  23.90   \\[0.1ex]
     chairs  (test) & \bf 24.26 &  17.17    \\[0.1ex]
    \end{tabular}
\end{center}    
\caption{Comparison of average PSNR metric between our model and autoencoder-based model~\cite{zimny2022points2nerf} trained on three classes of the same ShapeNet data. As we can see, we obtain slightly better renders.}
\label{tab:point}
\vspace{-0.3cm}
\end{wraptable}

The consequence of using non-trainable images is an explicit representation that uses fewer parameters than other hybrid methods, which need to have different data structures optimized during training. For evaluation, \our{} network contains similar architecture to NeRF~\cite{mildenhall2020nerf}, where the network uses $\sim$ 0.5M parameters. On the other hand, NSVF~\cite{liu2020neural} contains only one trainable network shared with other voxel grids. Although the render network is smaller, each scene requires a voxel-grid representation to be trained, which requires $\sim$ 3.2M parameters. 
In the case of the TriPlane model, it is difficult to give the number of parameters since TriPlane exists only as a part of a large model like GAN or diffiusions. 

\begin{wraptable}{r}{7.5cm}
\begin{center}
    \begin{tabular}{c|ccc}
    Trained &   & Render with &      \\[0.1ex]
    on &  cars &  chairs &  planes    \\
     \hline 
     cars & 24.91  &  22.15 & 21.32  \\[0.1ex] 
     chairs &  24.41 & 22.51 & 20.84 \\[0.1ex]
     planes & 24.19 &  21.69 & 24.27   \\[0.1ex]
    \end{tabular}
\end{center}
\caption{\our{} can generalize across different classes. The model is trained separately in three classes and evaluated in test sets from several classes. As we can see, the model gives similar results in training classes and unseen ones.}
\label{tab:unseen}
\end{wraptable}

\paragraph{\our{} for generalization}

Our model is a very small fully-connected architecture. In practice, we do not use 3D trainable representation. Consequently, we have less number of trainable parameters than other NeRF-based models.
Furthermore, such an approach allows us to generalize the model to unseen objects.

In practice, we need a full data set of 3D objects. Our experiments used a ShapeNet data set divided into training and test sets. Our model is trained on many objects with one category. After training, we fixed the weight of the implicit decoder. Then we use images from the test set without additional training, see Fig.~\ref{fig:generalization}. 

In such experiments, we add the cameras' positions to the input of an implicit decoder. In the case of training on one element, such positions do not change rendering results. Therefore we use camera positions only in generalization tasks.

When we built input to the implicit decoder, we applied projection on 2D images, concatenated colors and positions of projected points as well as potions of cameras 
$$
Z_{(x,y,z)} =  \left[ I_1[\mathbf{z}_{I_1}] , \mathbf{z}_{I_1} , P(I_1), \ldots, I_1[\mathbf{z}] , \mathbf{z}_{I_n}, P(I_n) \right] \in \R^{8 n},
$$
where $I_i[\mathbf{z}_{I_i}] \in \R^3$ is RGB color of position $\mathbf{z}_{I_i}$  on the image $I_i$, and $P(I_i)$ is position of camera dedicated to image $I_i$. 
Our implicit decoder $F_{MultiPlane NeRF}$ aggregates color and positions to produce  RGB colors and the volume density $\sigma$ 
$$
\F_{MultiPlane NeRF}(\x , {\bf d}; \Theta, I_1, P(I_1), \dots, I_n, P(I_n) ) = ( {\bf c} , \sigma). 
$$

The training procedure involves images from many different objects. The model is trained by sampling random renders from random objects. Each training batch has images of various objects taken from different positions.

\section{Experiments}

We evaluate the proposed \our{} on classical rendering and generalization tasks. In the first case, we compare our solution using the Diffuse Synthetic $360^{\circ}$ and Realistic Synthetic $360^{\circ}$ 3D object rendering provided by the original NeRF author's paper~\cite{mildenhall2020nerf}. In the generalization task, we use the data set dedicated to training the auto-encoder-based models Points2NeRF~\cite{zimny2022points2nerf}.

\begin{figure}
\begin{center}
\includegraphics[width=0.9\textwidth]{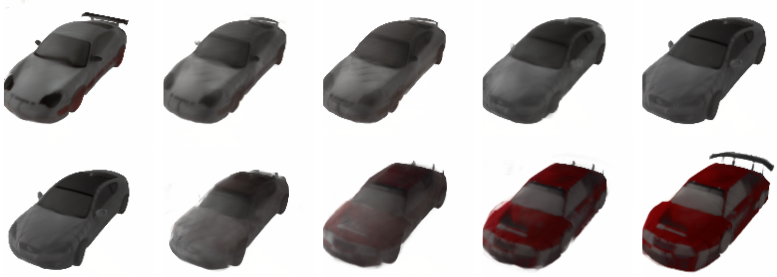}
\end{center}

\begin{wraptable}{r}{8.0cm}
\begin{center}
\begin{tabular}{lcccc}
\multirow{2}{*}{} & \multicolumn{2}{c}{FFHQ} & \multicolumn{2}{c}{Cars} \\ 

 & FID & KID & FID & KID \\ \hline

GIRAFFE $256^2$ & 31.5  & 1.992 & 27.3 & 1.703\\
$\pi$-GAN $128^2$ & 29.9 & 3.573 & 17.3 & 0.932 \\
Lift. SG $256^2$ & 29.8 & — & — & — \\
EG3D $128^2$ & — &  — & \bf 2.75 & \bf 0.097 \\
\ourGAN{} $128^2$ & — &  — & 6.4 & 0.309  \\
EG3D $512^2$ & \bf 4.7 & \bf 0.132 & — & — \\
\ourGAN{} $512^2$ & 15.4 &  1.007 & — & —  \\


\end{tabular}
\caption{
Quantitative evaluation using FID, KID×100, for FFHQ and ShapeNet Cars. 
}
\label{tab:gen_results}
\end{center}
\end{wraptable}

\paragraph{Synthetic renderings of objects} 

We first show experimental results on two
data sets of synthetic renderings of objects using the Diffuse Synthetic $360^{\circ}$ and Realistic Synthetic $360^{\circ}$. We compare our results with classical approaches SRN~\cite{sitzmann2019scene}, NV~\cite{lombardi2019neural},
LLFF~\cite{mildenhall2019local}, and original  NeRF~\cite{mildenhall2020nerf}, and Voxel based NSFV~\cite{liu2020neural}.

In Tab.~\ref{tab:nerfob}, we present a numerical comparison.
We compare the metric reported by NeRF called PSNR (\textit{peak signal-to-noise ratio}), SSIM (\textit{structural similarity index measure}), LPIPS (\textit{learned perceptual image patch similarity}) used to measure image reconstruction effectiveness.
We obtain similar results as vanilla NeRF and voxel-based NSFV, using fewer parameters. In Fig.~\ref{fig:rec_ner}, we present the qualitative results of \our{}.   

\our{} is a NeRF-based model with a non-trainable representation and mall implicit decoder, which obtain similar results as models with trainable representations and larger implicit decoder.

\begin{wrapfigure}{r}{7.5cm}
    \centering
    \qquad EG3D \qquad\qquad \ourGAN{} \\
    \includegraphics[width=0.11\textwidth, trim=15 15 15 15, clip]{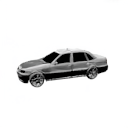}
    \includegraphics[width=0.11\textwidth, trim=15 15 15 15, clip]{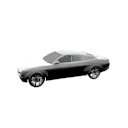}
    \includegraphics[width=0.11\textwidth, trim=7 7 7 7, clip]{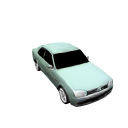}
    \includegraphics[width=0.11\textwidth, trim=7 7 7 7, clip]{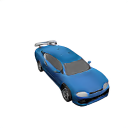}
    \includegraphics[width=0.11\textwidth, trim=15 15 15 15, clip]{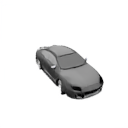}
    \includegraphics[width=0.11\textwidth, trim=15 15 15 15, clip]{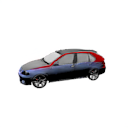}
    \includegraphics[width=0.11\textwidth, trim=7 7 7 7, clip]{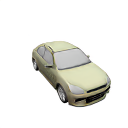}
    \includegraphics[width=0.11\textwidth, trim=7 7 7 7, clip]{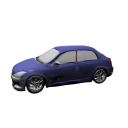}
    \includegraphics[width=0.11\textwidth, trim=15 15 15 15, clip]{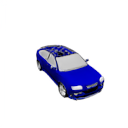}
    \includegraphics[width=0.11\textwidth, trim=15 15 15 15, clip]{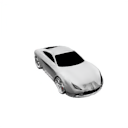}
    \includegraphics[width=0.11\textwidth, trim=7 7 7 7, clip]{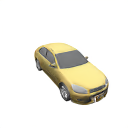}
    \includegraphics[width=0.11\textwidth, trim=7 7 7 7, clip]{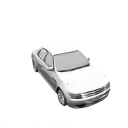}
    \caption{Comparison samples generated by \ourGAN{} and EG3D on ShapeNet Cars.}
    \label{fig:ing_ex_1}
    \vspace{-0.3cm}
\end{wrapfigure}

\paragraph{\our{} for generalization}

In the case of an experiment for generalization, we use a ShapeNet base data set containing 50 images of each element from the plane, chair, and car classes. 
For each object: fifty $200$x$200$ transparent background images from random camera positions. 
Such representation is perfect for training 3D models since each element has been seen from many views. The data was taken from \cite{zimny2022points2nerf}, where authors train an autoencoder-based generative model. 

In Fig.~\ref{fig:generalization}, we compare new renders obtained on the test set. As we can see, \our{} can generalize to unseen objects and obtain slightly better results than auto-encoder baser architecture. 
As we can see, we obtain good-quality objects, see Tab~\ref{tab:point}.

\caption{In \our{}, we use existing images as a non-trainable representation. In evaluation, we can render objects by mixing images from two objects. The figure shows the transition between objects   using $k$ images from the first object and $n-k$ from the second one. As we can see, our model produces reasonable interpolation between objects.   }
\label{fig:interpolation}
\end{figure}

Furthered more, our model can generalize across different classes. We train \our{} separately in three classes and evaluated in test sets from different classes. As we can see, the model gives similar results in training classes and unseen ones; see Tab.~\ref{tab:unseen}.

In \our{}, we use existing images as a non-trainable representation. In evaluation, we can render objects by mixing images from two objects. Fig~\ref{fig:interpolation} shows the transition between objects   using $k$ images from the first object and $n-k$ from the second one. We us $k=0\%, 20\%, 40\%, 60\% 80\%$ respectively. As we can see, our model produces reasonable interpolation between objects.

\paragraph{\our{} decoder in GAN architecture}

\begin{wrapfigure}{r}{7.5cm}
    \centering
    \qquad EG3D \qquad\qquad \ourGAN{} \\
    \includegraphics[width=0.1\textwidth]{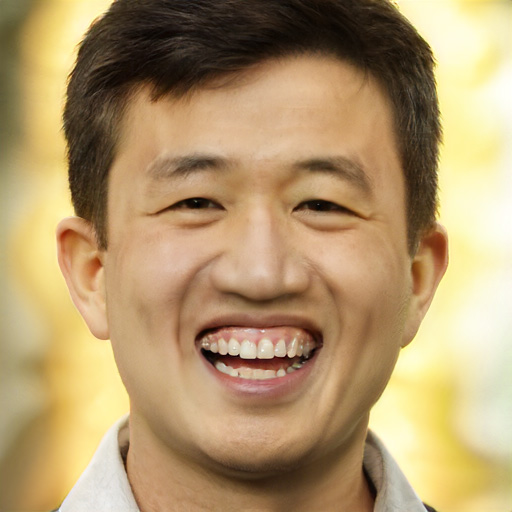}
    \includegraphics[width=0.1\textwidth]{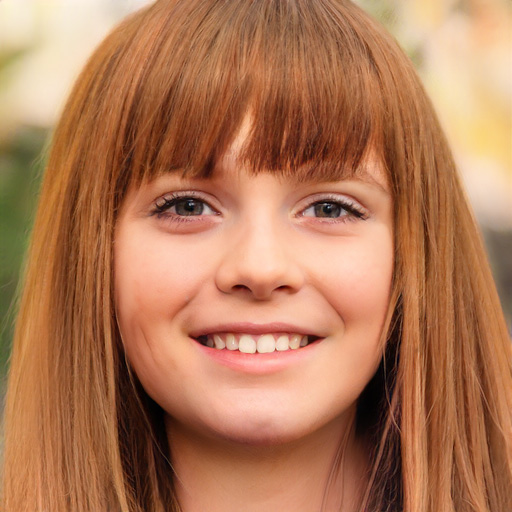} 
    \includegraphics[width=0.1\textwidth]{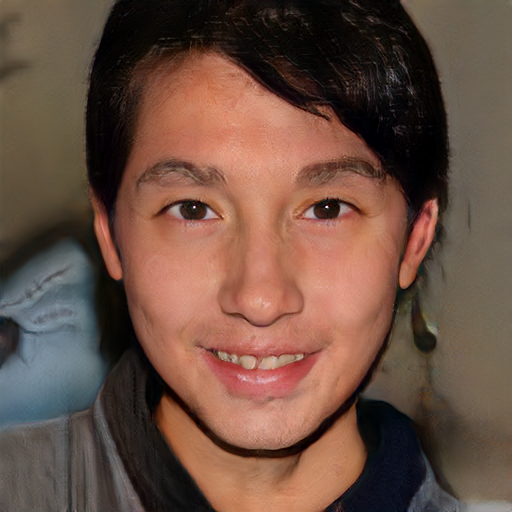} 
    \includegraphics[width=0.1\textwidth]{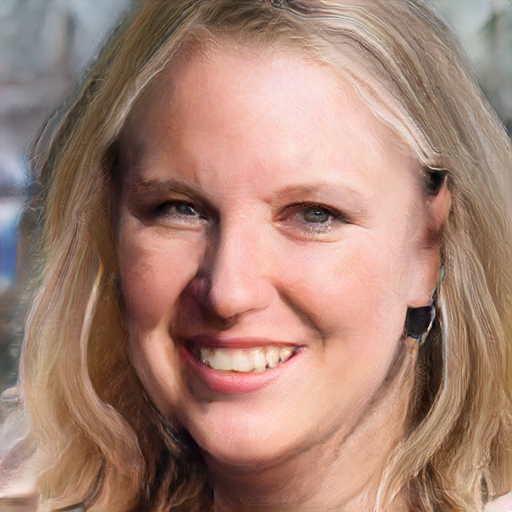}
    \\
    \includegraphics[width=0.1\textwidth]{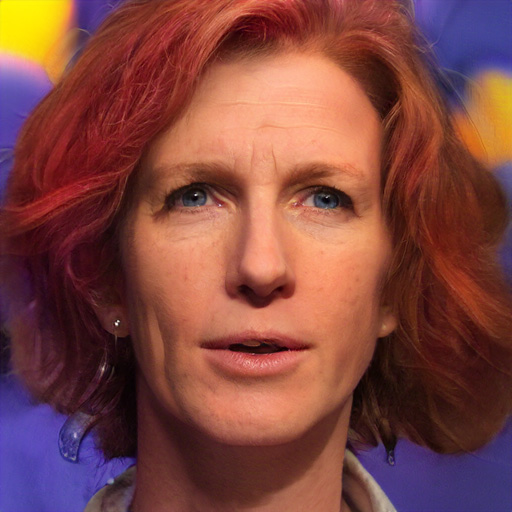}
    \includegraphics[width=0.1\textwidth]{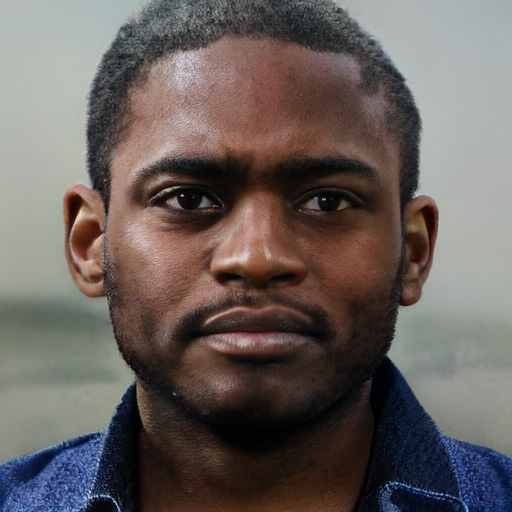}
    \includegraphics[width=0.1\textwidth]{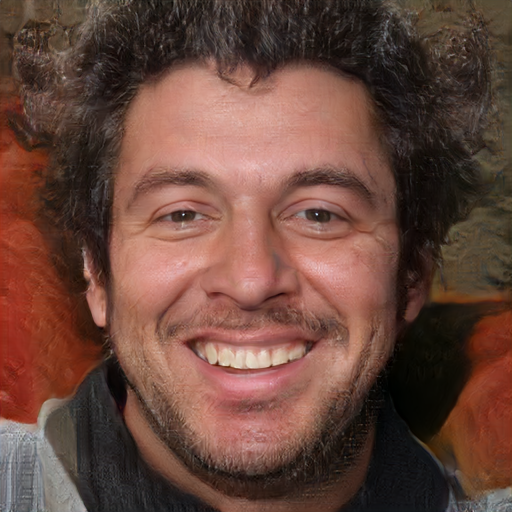}
    \includegraphics[width=0.1\textwidth]{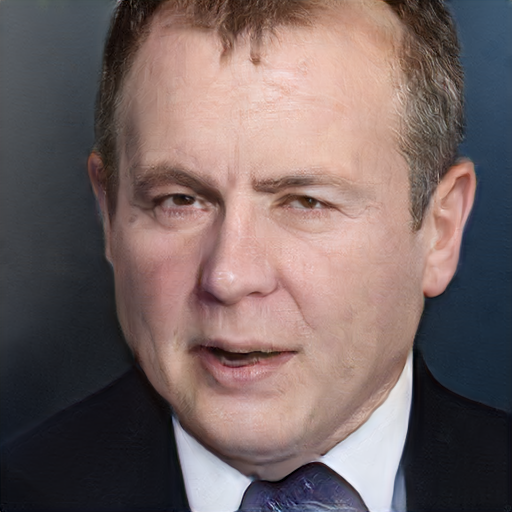}
     
    \caption{Comparison samples generated by \ourGAN{} and EG3D on FFHQ.}
    \label{fig:ing_ex_2}
\end{wrapfigure}

TriPlane decoder was introduced as part of the EG3D GAN~\cite{chan2022efficient}. EG3D uses a classical 2D generator to produce the TriPlane representation and 2D discriminator. Our MultiPlane decoder can be used as a part of larger architecture. We add a MultiPlane decoder to EG3D GAN to show such properties. 
As a result, we obtain MultiPlaneGAN, an analog of EG3D GAN~\cite{chan2022efficient} with MultiPlane decoder instead of TriPlane. In Tab.~\ref{tab:gen_results}, we compare MultiPlaneGAN with other models on two datasets FFHQ and ShapeNet Cars, see Fig.~\ref{fig:ing_ex_1} and Fig.~\ref{fig:ing_ex_2}. As we can see, we obtained the second-best score in both examples. We have slightly worse results than EG3D GAN, but we produce interpretable representation since our planes are 2D images with three RGB channels. 





\paragraph{Ablation studies} 
In \our{}, we use existing images as a non-trainable representation of 3D objects. To verify the influence of the number of such images and the resolution, we train \our{} on "Lego" model from NeRF dataset see Fig.~\ref{fig:abl1} and \ref{fig:abl2}. As we can see, our model obtains better results when we increase the number of images in representations. Also, the higher resolution allows us to obtain better-quality renders.

\section{Conclusions}
This paper presents a new NeRF model with easy-to-learn small architecture with generalization properties.
We use existing images instead of trainable representations like a voxel or TriPlane. We split the initial set of 2D training images into two subsets. The first one builds a 2D representation and is used as input to a small implicit decoder. The second part is used as a training data set to train the decoder. Using existing images as part of NeRF can significantly reduce the parameters since we train only a small implicit decoder. Furthermore, we can train \our{} on a large data set and force our implicit decoder to generalize across many objects. Therefore, we can only change the 2D image (without additional training) to generate NeRF representation of the new object.
\our{} gives comparable results to state-of-the-art models on synthesizing new views task. Moreover can generalize to unseen objects and classes. 

\paragraph{Limitations} The main limitation of the model is the trade-off between rendering quality and generalization properties. By training the model on a large dataset, we get a slightly worse quality than training each object separately. 



{\small
\bibliographystyle{plain}

}


\section{Appendix}

\begin{figure*}[!t]
    \centering
    \includegraphics[width=0.7\textwidth]{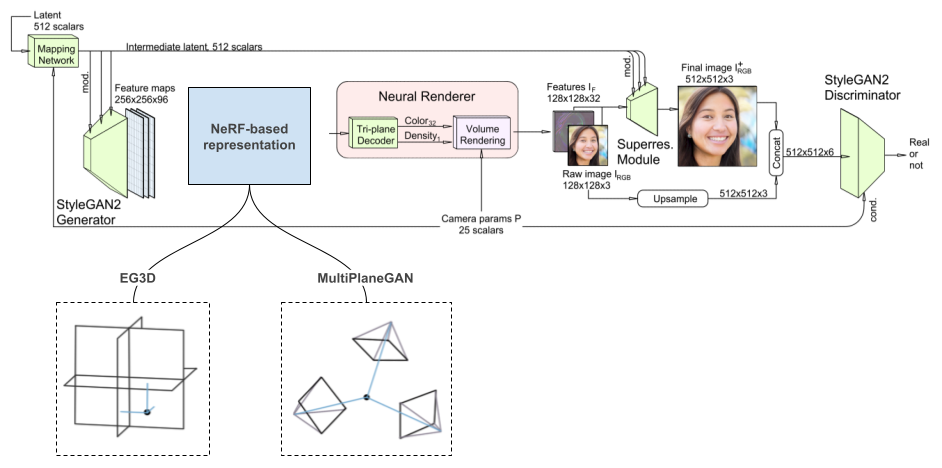}
        \caption{\ourGAN{} framework comprises several parts: a pose-conditioned StyleGAN2-based feature generator and mapping network, a MultiPlane 3D representation with a lightweight feature decoder, a neural volume renderer, a super-resolution module, and a poseconditioned StyleGAN2 discriminator with dual discrimination. Such architecture is based on EG3D GAN~\cite{chan2022efficient}. The main handicap of \ourGAN{} is using 2D image-based representations.
    }
    \label{fig:model}
\end{figure*}

MultiPlaneGAN is based on the EG3D GAN~\cite{chan2022efficient}. EG3D uses a classical 2D generator to produce the tri-plane representation and 2D discriminator. Our MultiPlane decoder can be used as a part of larger architecture. We add a MultiPlane decoder to EG3D GAN to show such properties. 
As a result, we obtain \ourGAN{}, an analog of EG3D GAN~\cite{chan2022efficient} with a MultiPlane decoder instead of tri-plane. 

In the EG3D we generate the tri-plane features containing 32 channels with the help of a 2D convolutional StyleGAN2
backbone. Instead of producing an RGB image, in the GAN setting EG3D neural renderer aggregates features from each of the 32-channel tri-planes and predicts 32-channel feature images from a given camera pose. This
is followed by a “super-resolution” module to upsample
and refine these raw neurally rendered images.
The generated images are critiqued by a slightly modified
StyleGAN2 discriminator (Sec. 4.3~\cite{chan2022efficient}).

\begin{wrapfigure}{r}{7.5cm}
    \centering
    \includegraphics[width=0.2\textwidth]{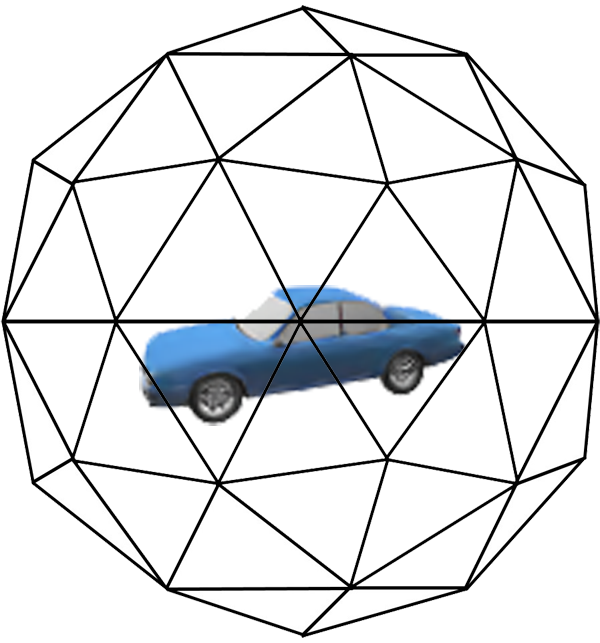}
     \caption{In \ourGAN{} we use MultiPlanes located on a sphere.}
    \label{fig:sphere}
\end{wrapfigure}

Fig. \ref{fig:model} gives an overview of \ourGAN{} architecture, which uses MultiPlane representation instead tri-plane. 
\ourGAN{} produces 32 planes consisting of 3 channels posed on a sphere containing an object.
To place planes in 3D space we use the icosphere, see Fig.~\ref{fig:sphere}.
Then MultiPlane decoder aggregates information to produce input to EG3D   super-resolution module and then to EG3D discriminator.

\paragraph{MultiPlaneGAN}

The entire pipeline is trained end-to-end from random initialization, using the non-saturating GAN loss function with R1 regularization, following the training scheme in StyleGAN2.

To speed training, EG3D use a two-stage training strategy in
which we train with a reduced (642) neural rendering resolution followed by a short fine-tuning period at full (1282) neural rendering resolution. Additional experiments found that regularization to encourage smoothness of the density
field helped reduce artifacts in 3D shapes. The following
sections discuss major components of our framework in detail. For additional descriptions, implementation details,
and hyperparameters, please see the supplement.

\end{document}